\documentclass{article}

\usepackage{graphicx}
\usepackage{amsmath}
\usepackage{amssymb}
\usepackage{amsthm}
\usepackage{natbib}
\usepackage{psfrag}

\newtheorem{theorem}{Theorem}

\newcommand{\bp}{\begin{pmatrix}}
\newcommand{\ep}{\end{pmatrix}}

\renewcommand{\b}{{\boldsymbol{b}}}

\let\x\undefined
\newcommand{\x}{{\boldsymbol{x}}}

\newcommand{\h}{{\boldsymbol{h}}}

\renewcommand{\r}{{\boldsymbol{r}}}

\renewcommand{\v}{{\boldsymbol{v}}}

\newcommand{\vbeta}{{\boldsymbol{\beta}}}

\newcommand{\veta}{{\boldsymbol{\eta}}}

\newcommand{\vtheta}{{\boldsymbol{\theta}}}

\DeclareMathOperator{\argmin}{argmin}

\DeclareMathOperator{\tr}{tr}

\begin{document}

\title{Approximating Incomplete Kernel Matrices by the {\it em} Algorithm}

\author{Koji Tsuda$^\dag$, Shotaro Akaho$^\ast$ and Kiyoshi Asai$^\dag$\\
       \qquad\\
	$^\dag$ AIST Computational Biology Research Center\\
	Tokyo, 135-0064, Japan\\
        $^\ast$ AIST Neuroscience Research Institute\\
        Tsukuba, 305-8568, Japan\\
{\tt \{koji.tsuda,s.akaho,asai-cbrc\}@aist.go.jp}
}

\maketitle

\begin{abstract}%
In biological data, it is often the case that observed data are
available only for a subset of samples.  When a kernel matrix is
derived from such data, we have to leave the entries for unavailable
samples as missing.  In this paper, we make use of a parametric model
of kernel matrices, and estimate missing entries by fitting the model
to existing entries.  The parametric model is created as a set of
spectral variants of a complete kernel matrix derived from another
information source.  For model fitting, we adopt the {\it em}
algorithm based on the information geometry of positive definite
matrices.  We will report promising results on bacteria clustering
experiments using two marker sequences: 16S and gyrB.
\end{abstract}

\section{Introduction}
In kernel machines such as support vector machines (SVM)~\citep{SchSmo01},
objects are represented as a kernel matrix, 
where $n$ objects are represented as 
an $n \times n$ positive semidefinite matrix.
Essentially the $(i,j)$ entry of the kernel matrix describes 
the similarity between $i$-th and $j$-th objects.
Due to positive semidefiniteness, 
the objects can be embedded as $n$ points in an Euclidean feature space 
such that the inner product between two points equals to 
the corresponding entry of kernel matrix.
This property enables us to apply diverse learning methods 
(for example, SVM or kernel PCA) without explicitly constructing 
a feature space~\citep{SchSmo01}.
 
Biological data such as amino acid sequences, gene expression arrays
and phylogenetic profiles are derived from expensive experiments~\citep{Bro99}.
Typically initial experimental measurements are so noisy 
that they cannot be given to learning machines directly.
Since high quality data are created by extensive work of human experts,
it is often the case that good data 
are available only for a subset of samples.
When a kernel matrix is derived from such incomplete data,
we have to leave the entries for unavailable samples as {\it missing}.
We call such a matrix an ``incomplete matrix''.
Our aim is to estimate the missing entries, 
but it is obviously impossible without additional information.
So we make use of a {\it parametric model} of admissible matrices, 
and estimate missing entries by fitting the model to existing entries.

In this scheme, it is important to define a parametric model appropriately.
For example, \citet{Gra02} used the set of all positive definite matrices 
as a model.
Although this model worked well when only a few entries are missing,
this model is too general for our cases where whole columns and rows 
are missing.
Thus we need another information source for constructing a parametric model.
Fortunately, in biological data, it is common that one object 
is described by two or more representations.
For example, genes are represented by gene networks 
and gene expression arrays at the same time~\citep{VerKan02}.
Also a bacterium is represented by several marker 
sequences~\citep{YamKasArnJacVivHar00}.
In this paper, we assume that a complete matrix is available 
from another information source, and a parametric model is created
by giving perturbations to the matrix.
We call the complete matrix a ``base matrix''.
When creating a parametic model of admissible matrices from a base matrix,
one typical way is to define the parametric model 
as all {\it spectral variants} of the base matrix, 
which have the same eigenvectors but different 
eigenvalues~\citep{CriShaKanEli02}.
When several base matrices are available,
the weighted sum of these matrices would be a good parametric model 
as well~\citep{LanCriBarElGJor02}. 

In order to fit a parametric model,
the distance between two matrices has to be determined.
A common way is to define the Euclidean distance between matrices 
(for example, the Frobeneous norm) and make use of the Euclidean geometry.
Recently \citet{VerKan02} tackled with the incomplete matrix 
approximation problem by means of kernel CCA.
Also \citet{CriShaKanEli02} proposed a similarity measure called ``alignment'',
which is basically the cosine between two matrices.
In contrast that their methods are based on the Euclidean geometry,
this paper will follow an alternative way: we will define 
the Kullback-Leibler (KL) divergence between two kernel matrices 
and make use of the Riemannian information geometry~\citep{OhaSudAma96}.
The KL divergence is derived by relating a kernel matrix 
to a covariance matrix of  Gaussian distribution.
The primal advantage is that the KL divergence allows us to use 
the $em$ algorithm~\citep{Ama95} to approximate an incomplete kernel matrix.
The $e$ and $m$ steps are formulated 
as convex programming problems, and moreover they can be solved analytically 
when spectral variants are used as a parametric model.

We performed bacteria clustering experiments 
using two marker sequences: 16S and gyrB~\citep{YamKasArnJacVivHar00}.
We derived the incomplete and base kernel matrices 
from gyrB and 16S, respectively.
As a result, even when 50\% of columns/rows are missing,
the clustering performance of the completed matrix
was better than that of the base matrix,
which illustrates the effectiveness of our approach in real world problems.

This paper is organized as follows: 
Sec.~\ref{sec:ig} introduces the information geometry to the space
of positive definite matrices.
Based on geometric concepts, the {\it em} algorithm for matrix
approximation is presented in Sec.~\ref{sec:em}, 
where detailed computations are deferred in Sec.~\ref{sec:compproj}.
In Sec.~\ref{sec:EM}, the matrix approximation problem is formulated
as statistical inference and the equivalence between the {\it em} and 
EM algorithms~\citep{DemLaiRub77} is shown.
Then the bacteria clustering experiment is described 
in Sec.~\ref{sec:experiment}.
After seeking for possible extensions in Sec.~\ref{sec:ext},
we conclude the paper in Sec.~\ref{sec:con}.

\section{Information Geometry of Positive Definite Matrices} \label{sec:ig}
We first explain how to introduce the information geometry 
in the space of positive definite matrices.
Only necessary parts of the theory will be presented here, 
so refer to \citep{OhaSudAma96,AmaNag01} for details.

Let us define the set of all $d \times d$ positive definite matrices 
as ${\cal P}$.
The first step is to relate a $d \times d$ positive definite matrix 
$P \in {\cal P}$ 
to the Gaussian distribution with mean 0 and covariance matrix $P$:
\begin{equation} \label{eq:Gaussian}
p(\x | P) = 
\frac{1}{(2 \pi)^{d/2} |P|^{1/2}} \exp (-\frac{1}{2} \x^\top P^{-1} \x).
\end{equation}
It is well known that the Gaussian distribution belongs to 
the exponential family. 
The canonical form of an exponential family distribution is written as
\[
p(\x | \vtheta) = \exp( \vtheta^\top \r(\x) - \psi(\vtheta)),
\]
where $\r(\x)$ is the vector of sufficient statistics, 
$\vtheta$ is the natural parameter and $\psi(\vtheta)$ 
is the normalization factor.
When (\ref{eq:Gaussian}) is rewritten in the canonical form, 
we have the sufficient statistics as
\[
\r(\x) = - \left( \frac{1}{2} x_1^2, \ldots, \frac{1}{2} x_d^2, 
x_1 x_2, \ldots, x_{d-1} x_d \right)^\top,
\]
and the natural parameter as
\[
\vtheta = \left( [P^{-1}]_{11}, \ldots, [P^{-1}]_{dd}, 
[P^{-1}]_{12}, \ldots, [P^{-1}]_{d-1,d} \right)^\top,
\]
where $[M]_{ij}$ denotes the $(i,j)$ entry of matrix $M$.
The natural parameter $\vtheta$ provides a coordinate system 
to specify a positive definite matrix $P$, 
which is called the $\vtheta$-coordinate system (or the $e$-coordinate system).
On the other hand, there is an alternative representation 
for the exponential family.
Let us define the mean of $r_i(x)$ as $\eta_i$:
For example, when $r_i(x) = x_s x_t$,
\[
\eta_i = \int x_s x_t p(\x | \vtheta) d\x = P_{st}.
\]
This new set of parameters $\eta_i$ provides another coorninate system,
called $\veta$-coordinate system (or the $m$-coordinate system):
\[
\veta = \left( P_{11}, \ldots, P_{dd}, P_{12}, \ldots, P_{d-1,d} \right)^\top.
\]
Let us consider the following curve $\vtheta(t)$ connecting two points 
$\vtheta_1$ and $\vtheta_2$ linearly in $\vtheta$ coordinates:
\[
\vtheta(t) = t(\vtheta_2 - \vtheta_1) + \vtheta_1.
\]
When written is the matrix form, this reads
\[
P^{-1}(t) = t(P^{-1}_2 - P^{-1}_1) + P^{-1}_1.
\]
This curve is regarded as a straight line from the exponential viewpoint
and is called an exponential geodesic or $e$-geodesic.
In particular, each coordinate curve $\theta_i = t$, 
$\theta_j = c_j~~ (j \neq i)$ is an $e$-geodesic.
When the $e$-geodesic between any two points in a manifold
${\cal S} \subseteq {\cal P}$ 
is included in ${\cal S}$, the manifold ${\cal S}$ 
is said to be $e$-flat.
On the other hand, the mixture geodesic or $m$-geodesic is defined as
\[
\veta(t) = t(\veta_2 - \veta_1) + \veta_1.
\]
In the matrix form, this reads
\[
P(t) = t(P_2 - P_1) + P_1.
\]
When the $m$-geodesic between any two points in ${\cal S}$ 
is included in ${\cal S}$, the manifold ${\cal S}$ is said to be $m$-flat.


In information geometry,
the distance between probability distributions is defined as 
the Kullback-Leibler divergence~\citep{AmaNag01}:
\[
KL(p,q) = \int p(x) \log \frac{p(x)}{q(x)} dx.
\]
By relating a positive definite matrix 
to the covariance matrix of Gaussian (\ref{eq:Gaussian}),
we have the Kullback-Leibler (KL) divergence for two matrices $P, Q$:
\[
KL(P,Q) = \tr(Q^{-1} P) + \log \det Q - \log \det P - d.
\]
With respect to a manifold ${\cal S} \subseteq {\cal P}$ and 
a point $P \in {\cal P}$,
the projection from $P$ to ${\cal S}$ is defined as the point 
in ${\cal S}$ closest to $P$.
Since the KL divergence is asymmetric, there are two kinds of projection:
\begin{itemize}
\item $e$-projection: $Q^* = \argmin_{Q \in {\cal S}} KL(Q,P)$.
\item $m$-projection: $Q^* = \argmin_{Q \in {\cal S}} KL(P,Q)$.
\end{itemize}
It is proved that the $m$-projection to an $e$-flat submanifold is unique, 
and $e$-projection to an $m$-flat manifold is unique~\citep{AmaNag01}.
This uniqueness property means that 
the corresponding optimization problem is convex and 
so the global optimal solution is easily obtained by any reasonable method.


\section{Approximating an Incomplete Kernel Matrix} \label{sec:em}
In this section, we describe the {\it em} algorithm to approximate 
an incomplete kernel matrix.
Let $x_1,\ldots,x_\ell \in {\cal X}$ be the set of samples in interest.
In supervised learning cases, this set includes both training and test sets,
thus we are considering the transductive setting~\citep{Vap98}.
Let us assume that the data is available for the first $n$ samples,
and unavailable for the remaining $m := \ell-n$ samples.
Denote by $K_I$ an $n \times n$ kernel matrix, 
which is derived from the data for the first $n$ samples.
Then, an incomplete kernel matrix is described as
\begin{equation} \label{eq:D}
D = \left( \begin{array}{cc} K_I & D_{vh} \\ D_{vh}^\top & D_{hh} 
\end{array} \right),
\end{equation}
where $D_{vh}$ is an $n \times m$ matrix and 
$D_{hh}$ is an $m \times m$ symmetric matrix.
Since $D$ has missing entries, 
it cannot be presented as a point in ${\cal P}$.
Instead, all the possible kernel matrices form a manifold
\[
{\cal D} = \{ D \; | \; D_{vh} \in \Re^{n \times m}, \;\;  D_{hh} \in \Re^{m \times m}, \;\;  D_{hh} = D_{hh}^\top,  \;\; D \succ 0  \},
\]
where $D \succ 0$ means that $D$ is positive definite.
We call it the {\it data manifold} 
as in the conventional EM algorithm~\citep{IkeAmaNak99}.
It is easy to verify that $\cal D$ is
an $m$-flat manifold; hence, the $e$-projection
to $\cal D$ is unique.

Next let us define the parametric model to approximate $D$.
Here the model is derived as the spectral variants of $K_B$, 
which is an $\ell \times \ell$ base kernel matrix derived from
another information source.
Let us decompose $K_B$ as
\[
K_B = \sum_{i=1}^\ell \lambda_i \v_i \v_i^\top,
\]
where $\lambda_i$ and $\v_i$ is the $i$-th eigenvalue and eigenvector,
respectively. Define 
\begin{equation} \label{eq:mi}
M_i = \v_i \v_i^\top,
\end{equation} 
then all the spectral variants are represented as
\[
{\cal M} = \{ M \; | \; M = \sum_{j=1}^\ell \beta_j M_j, \;\; \vbeta \in \Re^\ell, \;\; M \succ 0 \}
\]
We call it the model manifold~\citep{IkeAmaNak99}.
For notational simplicity, we choose a different parametrization 
of ${\cal M}$:
\begin{equation} \label{eq:mdef}
{\cal M} = \{ M \; | \; M = ( \sum_{j=1}^\ell b_j M_j )^{-1}, \;\; \b \in \Re^\ell, \;\; M \succ 0 \},
\end{equation}
where $b_j = 1/\beta_j$.
It is easily seen that the manifold ${\cal M}$ 
is $e$-flat and $m$-flat at the same time. 
Such a manifold is called dually-flat.

\begin{figure}[t]
\psfrag{ddd}{${\cal D}$}
\psfrag{mmm}{${\cal M}$}
\begin{center}
\includegraphics[width=0.5\textwidth]{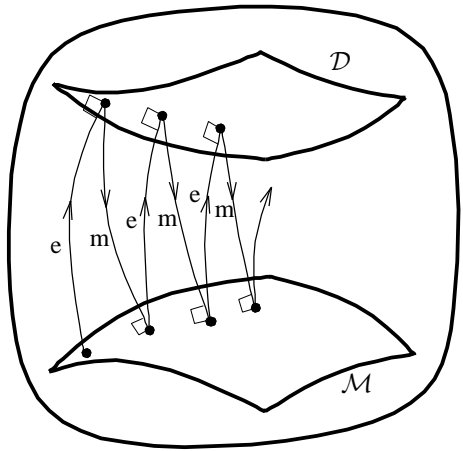}
\end{center}
\caption{Information geometric picture of the $em$ algorithm.
The data manifold ${\cal D}$ corresponds to the set of 
all completed matrices, whereas the model manifold ${\cal M}$ corresponds
to the set of all spectral variants of a base matrix. 
The nearest points are found by gradually minimizing the KL divergence
by repeating $e$ and $m$ projections.}
\label{fig:em}
\end{figure}

Our approximation problem is formulated as finding the nearest points 
in two manifolds: Find $D \in {\cal D}$ and $M \in {\cal M}$ 
to minimize $KL(D,M)$.
In geometric terms, this problem is to find the nearest points 
between $e$-flat and $m$-flat manifolds.
It is well known that such a problem is solved by an alternating procedure
called the {\it em} algorithm~\citep{Ama95}.
The {\it em} algorithm gradually minimizes the KL divergence 
by repeating $e$-step and $m$-step alternately (Fig.~\ref{fig:em}).

In the $e$-step, the following optimization problem is solved with fixing $M$:
Find $D \in {\cal D}$ that minimizes $KL(D,M)$.
This is rewritten as follows: 
Find $D_{vh}$ and $D_{hh}$ that minimize
\begin{equation} \label{eq:eopt}
L_e = \tr(D M^{-1}) - \log \det D,
\end{equation}
subject to the constraint that $D \succ 0$.
Notice that this constraint is not needed, because
\[
\log \det D = \sum_{i=1}^\ell \log \mu_i,
\]
where $\mu_i$ is the $i$-th eigenvalue of $D$.
Here $\log \det D$ is undefined when one of eigenvalues is negative, 
and $\log \det D$ decreases to $-\infty$ as an eigenvalue get closer to 0.
So, at the optimal solution, $D$ is necessarily positive definite,
because the KL divergence is infinite otherwise.
As indicated by information geometry, this is a convex problem, 
which can readily be solved by any reasonable optimizer. 
Moreover the solution is obtained in a closed form:
Let us partition $M^{-1}$ as
\begin{equation} \label{eq:mdivide}
M^{-1} = \left( \begin{array}{cc} S_{vv} & S_{vh} \\
S_{vh}^\top & S_{hh} \end{array} \right).
\end{equation}
The solution of (\ref{eq:eopt}) is described as
\begin{eqnarray} 
D_{vh} &=& - K_I S_{vh} S_{hh}^{-1}, \label{eq:dvhlast} \\
D_{hh} &=& S_{hh}^{-1} + S_{hh}^{-1} S_{vh} K_I S_{vh} S_{hh}^{-1}. \label{eq:dhhlast}
\end{eqnarray}
The derivation of (\ref{eq:dvhlast}) and (\ref{eq:dhhlast}) 
will be described in Sec.~\ref{sec:eproj}.

In the $m$-step, the following optimization problem is solved with fixing $D$:
Find $M \in {\cal M}$ that minimizes $KL(D,M)$.
This is rewritten as follows:
Find $\b \in \Re^\ell$ that minimizes
\begin{equation} \label{eq:mopt}
L_m = \sum_{j=1}^\ell b_j \tr(M_j D)  - \log \det (\sum_{j=1}^\ell b_j M_j)
\end{equation}
subject to the constraint that $\sum_{j=1}^\ell b_j M_j \succ 0$.
Notice that this constraint can be ignored as well.
When $\{M_j\}_{j=1}^\ell$ are defined as (\ref{eq:mi}),
the closed form solution of (\ref{eq:mopt}) is obtained as
\begin{equation} \label{eq:msol}
b_i = 1 / \tr(M_i D), \qquad i = 1,\ldots,\ell.
\end{equation}
The derivation of (\ref{eq:msol}) will be described in Sec.~\ref{sec:mproj}.

\section{Computing Projections} \label{sec:compproj}
This section presents the derivation of $e$ and $m$-projections in detail.

\subsection{$e$-projection} \label{sec:eproj}
First we will show the derivation of $e$-projection 
(\ref{eq:dvhlast}) and (\ref{eq:dhhlast}).
The log determinant of a partitioned matrix is rewritten as 
\begin{eqnarray*}
L_e &=& \tr(D M^{-1}) - \log \det D \\
&=& \tr(D M^{-1}) 
- \log \det K_I - 
\log \det (D_{hh} - D_{vh}^\top K_I^{-1} D_{vh}).
\end{eqnarray*}
When we partition $M^{-1}$ as (\ref{eq:mdivide}),
it turns out that
\begin{equation}\label{eq:le}
L_e = \tr(D_{vv} S_{vv}) + 2 \tr(D_{vh} S_{vh}) + \tr(D_{hh} S_{hh})
- \log \det K_I - 
\log \det (D_{hh} - D_{vh}^\top K_I^{-1} D_{vh}).
\end{equation}
The saddle point equation with respect to $D_{hh}$ is obtained as
\begin{equation} \label{eq:dhh}
\frac{\partial L_e}{\partial D_{hh}} = 
S_{hh} - (D_{hh} - D_{vh}^\top K_I^{-1} D_{vh})^{-1},
\end{equation}
because $\frac{\partial}{\partial C} \log \det C = C^{-1}$ 
for any symmetric matrix $C$.
Solving (\ref{eq:dhh}) with respect to $D_{hh}$, we have
\begin{equation} \label{eq:dhhmid}
D_{hh} = S_{hh}^{-1} + D_{vh}^\top K_I^{-1} D_{vh}.
\end{equation}
Substituting (\ref{eq:dhhmid}) into (\ref{eq:le}), we have
\[
L_e = \tr(D_{vv} S_{vv}) + 2 \tr(D_{vh} S_{vh}) + \tr(I + S_{hh} D_{vh}^\top K_I^{-1} D_{vh}) - \log \det K_I - \log \det S_{hh}.
\]
Now the saddle point equation with respect to $D_{vh}$ is obtained as
\[
\frac{\partial L_e}{\partial D_{vh}} = 
2 S_{vh} + 2 K_I^{-1} D_{vh} S_{hh} = 0.
\]
Solving this equation, we have the solution (\ref{eq:dvhlast}) for $D_{vh}$.
By substituting (\ref{eq:dvhlast}) into (\ref{eq:dhhmid}), 
we have the solution (\ref{eq:dhhlast}) for $D_{hh}$.

\subsection{$m$-projection} \label{sec:mproj}
Next, we will show the derivation of $m$-projection (\ref{eq:msol}).
The $m$ projection is obtained as the solution $\b$ to minimize
\[
L_m = \sum_{j=1}^\ell b_j \tr(M_j D) - \log\det(\sum_{j=1}^\ell b_j M_j).
\]
Since $\partial\log\det Q^{-1}/\partial Q = Q$, 
the saddle point equations are described as
\begin{equation}
\label{eq:mstep}
 \tr(M_i (\sum_{j=1}^c b_j M_j)^{-1}) = \tr(M_i D),
 \qquad i = 1,\ldots,\ell.
\end{equation}
Remembering that $M_j = \v_j \v_j^\top$, we have
\[
(\sum_{j=1}^\ell b_j M_j)^{-1} = \sum_{j=1}^\ell \frac{1}{b_j} \v_j \v_j^\top.
\]
Since the left hand side of (\ref{eq:mstep}) is 
\[
\tr(\v_i \v_i^\top \sum_{j=1}^\ell \frac{1}{b_j} \v_j \v_j^\top)
= \tr (\frac{1}{b_i} \v_i \v_i^\top) = \frac{1}{b_i},
\]
the solution of (\ref{eq:mstep}) is analytically obtained as
\[
b_i = 1 / \tr(M_i D), \qquad i = 1,\ldots,\ell.
\]
We have shown that the $m$-projection is obtained analytically 
when the model manifold corresponds to spectral variants of a matrix.
However, it is not always the case. 
For example, consider we have $c$ base matrices $N_1,\ldots,N_c$ and 
the model manifold is constructed as harmonic mixture of them:
\begin{equation} \label{eq:generalM}
{\cal M} = \{ M \; | \; M = ( \sum_{j=1}^c b_j N_j )^{-1}, \;\; \b \in \Re^c, \;\; M \succ 0 \}.
\end{equation}
This is an $e$-flat manifold so the optimization problem is convex, 
but the analytical solvability depends on geometric properties 
of base matrices $\{N_i\}_{i=1}^c$\citep{Oha98}. 
We will briefly discuss this issue in the Appendix.

\section{Relation to the EM algorithm} \label{sec:EM}
In statistical inference with missing data,
the EM algorithm~\citep{DemLaiRub77} is commonly used.
By posing the matrix approximation problem as statistical inference,
the EM algorithm can be applied, and --- as shown later --- 
it eventually leads to the same procedure.
In a sense, it is misleading to relate matrix approximation 
to statistical concepts, such as random variables, observations and so on.
Nevertheless it would be meaningful to rewrite our method in terms of 
statistical concepts for establishing connections to other literature. 

Let $\v$ and $\h$ be the $n$ and $m$ dimensional visible 
and hidden variables.
From observed data\footnote{In fact, we do not have observed data 
in any sense. However, we assumed them as a matter of form 
for relating $em$ and $EM$.}, 
the covariance matrix of $\v$ is known as
\[
E_o[\v \v^\top] = K_I,
\]
where $E_o$ denotes the expectation with respect to observed data.
However, we do not know the covariances 
$D_{vh} = E_o[\v \h^\top]$ and $D_{hh} = E_o[\h \h^\top]$.
Our purpose is to obtain the maximum likelihood estimate of 
parameter $\b$ of the following Gaussian model:
\[
p(\v,\h | \b) = 
\frac{1}{(2 \pi)^{d/2} |M|^{1/2}}
\exp \left( -\frac{1}{2} \left[ \begin{array}{c} \v \\ \h \end{array} \right]^\top M^{-1} \left[ \begin{array}{c} \v \\ \h \end{array} \right] \right),
\]
where $M$ is described as (\ref{eq:mdef}).
In the course of maximum likelihood estimation,
we have to estimate the observed covariances 
$D_{vh}$ and $D_{hh}$ in an appropriate way.
The EM algorithm consists of the following two steps.
\begin{itemize}
\item E-Step: Fix $\b$ and update $D_{vh}$ and $D_{hh}$ 
by conditional expectation.
\item M-Step: Fix $D$ and update $\b$ by maximum likelihood estimation.
\end{itemize}
It is shown that the likelihood of observed data increases 
monotonically by repeating these two steps~\citep{DemLaiRub77}.

The M-step maximizes the likelihood, 
which is easily seen to be equivalent 
to minimizing the KL divergence~\citep{Ama95}.
So the $M$-step is equivalent to the $m$-step (\ref{eq:mopt}).
However, the equivalence between E-step and $e$-step is not obvious,
because the former is based on conditional expectation 
and the latter minimizes the KL divergence.
In the E-step, the covariance matrices are computed
from the conditional distribution described as
\[
p(\h|\v,\b) = \frac{1}{(2 \pi)^{m/2} | S_{hh}^{-1} |^{1/2}}
\exp \left( - \frac{1}{2} (\h + S_{hh}^{-1} S_{vh}^\top \v)^\top
S_{hh} (\h + S_{hh}^{-1} S_{vh}^\top \v) \right),
\]
where $S$ matrices are derived as (\ref{eq:mdivide}).
Taking expectation with this distribution, we have 
\begin{eqnarray*}
E_\b[\v\h^\top\mid\v] &=& -\v\v^\top S_{vh} S_{hh}^{-1}, \\
E_\b[\h\h^\top\mid\v] &=& S_{hh}^{-1} + S_{hh}^{-1}S_{vh}^\top\v\v^\top S_{vh} S_{hh}^{-1}.
\end{eqnarray*}
Then the covariance matrices are estimated as
\begin{eqnarray*}
D_{vh} = E_o E_\b[\v\h^\top\mid\v] &=& - K_I S_{vh} S_{hh}^{-1}, \\
D_{hh} = E_o E_\b[\h\h^\top\mid\v] &=& S_{hh}^{-1} + S_{hh}^{-1}S_{vh}^\top K_I S_{vh} S_{hh}^{-1}.
\end{eqnarray*}
Since these solutions are equivalent to (\ref{eq:dvhlast}) and (\ref{eq:dhhlast}), respectively, 
the E-step is shown to be equivalent to the $e$-step in this case.
Refer to \citet{Ama95} for general discussion of 
the equivalence between EM and $em$ algorithms.

\section{Bacteria Classification Experiment} \label{sec:experiment}
In this section, we perform unsupervised classification experiments
for bacteria based on two marker sequences: 16S and gyrB.
Basically we would like to identify the genus of a bacterium 
by means of extracted entities from the cell.
It is known that several specific proteins and RNAs 
can be used for genus identification~\citep{KasEtal98}.
Among them, we especially focus on 
16S rRNA and gyrase subunit B (gyrB) protein.
16S rRNA is an essential constituent in all living organisms,
and the existence of many conserved regions in the rRNA genes allows
the alignment of their sequences derived from distantly related organisms,
while their variable regions are useful for the distinction of closely
related organisms.
GyrB is a type II DNA topoisomerase which is an enzyme that controls
and modifies the topological states of DNA supercoils.
This protein is known to be well preserved over evolutional history
among bacterial organisms thus is supposed to be a better identifier
than the traditional 16S rRNA \citep{KasEtal98}.
Notice that 16S is represented as a nucleotide sequence with 4 symbols, 
and gyrB is an amino acid sequence with 20 symbols.
Since gyrB has been found to be useful more recently 
than 16S~\citep{YamKasArnJacVivHar00}, 
gyrB sequences are available only for a limited number of bacteria.
Thus, it is considered that gyrB is more ``expensive'' than 16S.

\begin{figure}
\begin{center}
\begin{tabular}{cc}
(a) Ideal & (b) gyrB (complete) \\
\includegraphics[width=0.4\textwidth]{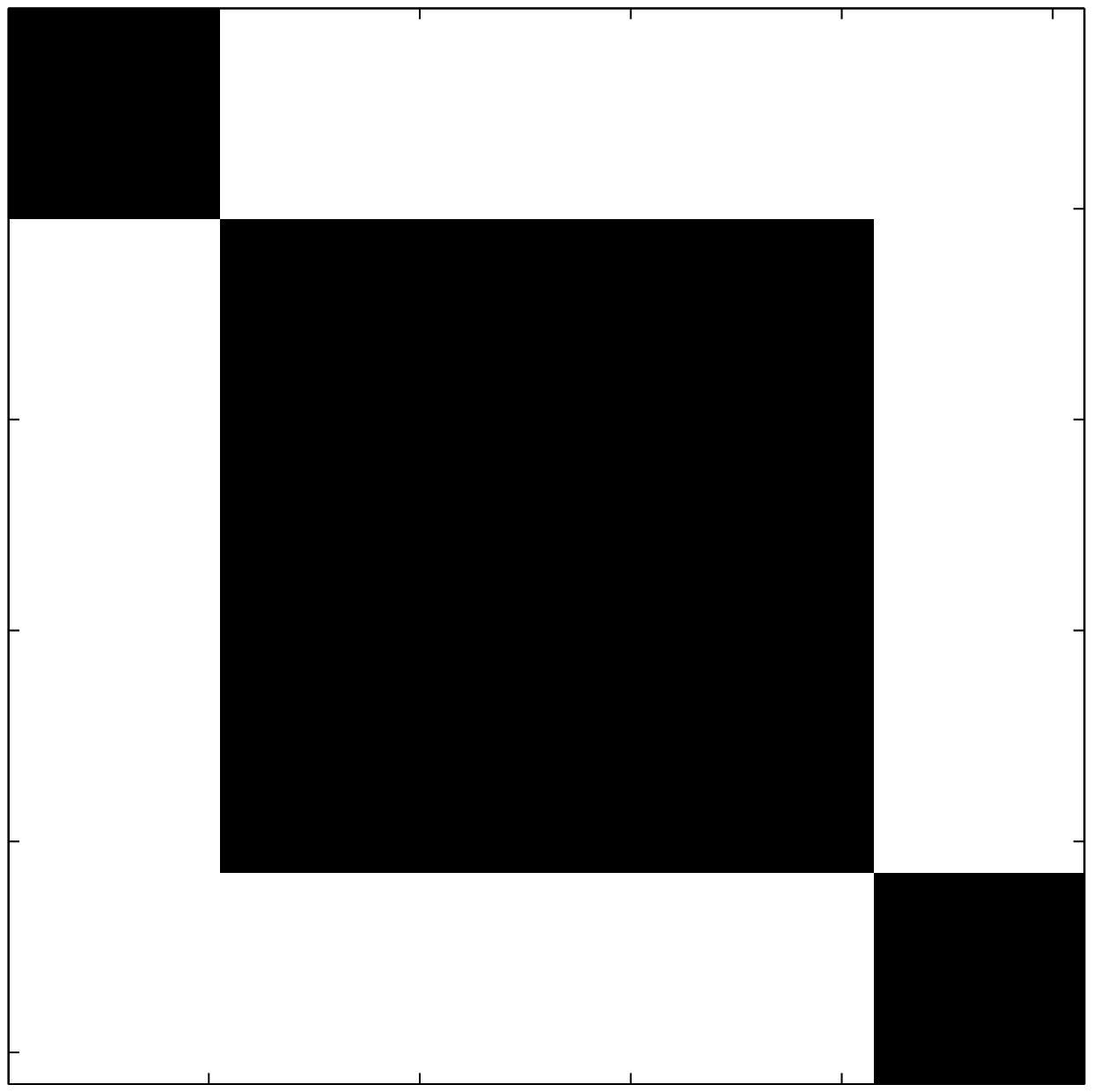} &
\includegraphics[width=0.4\textwidth]{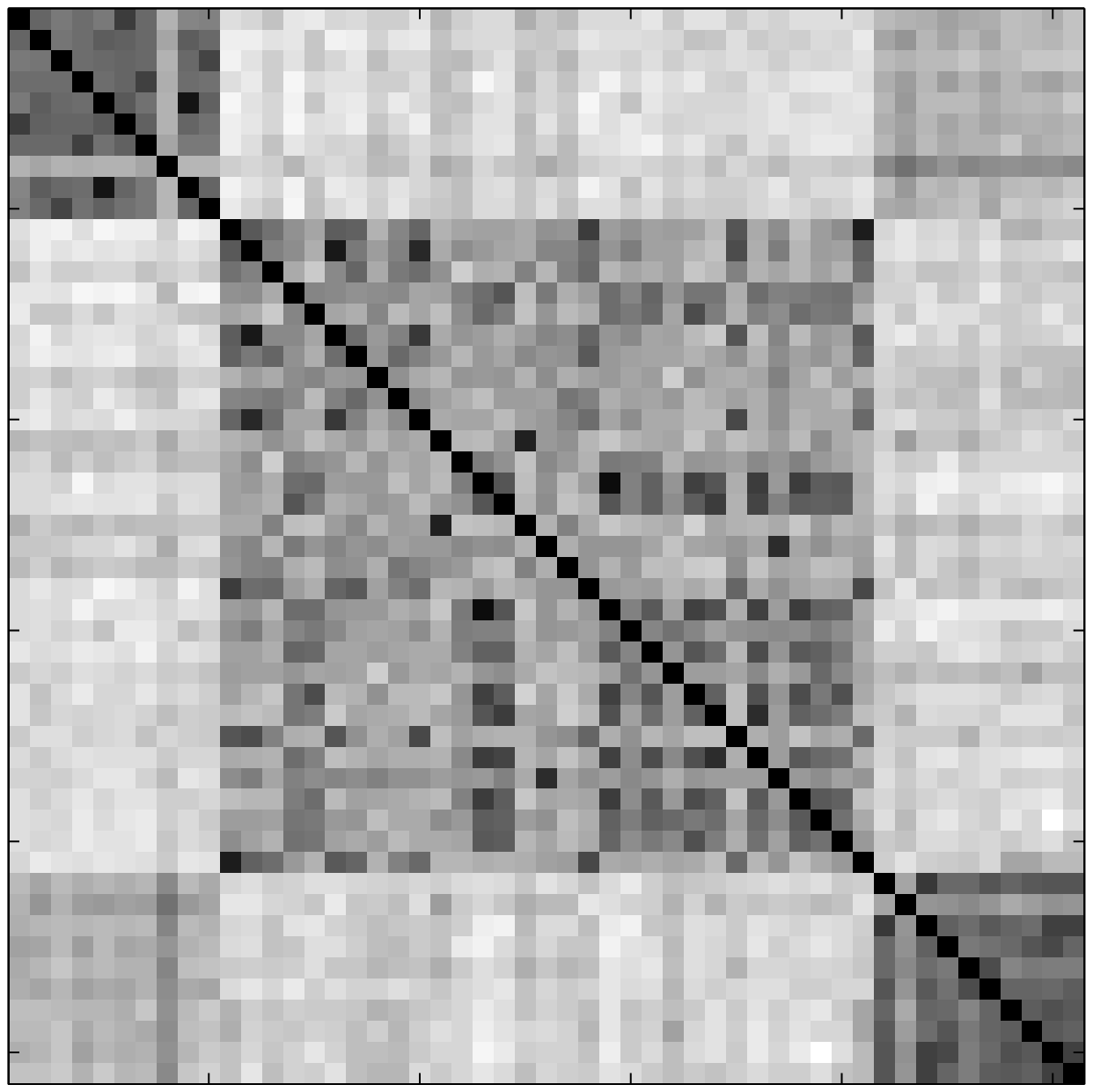} \\
(c) 16S & (d) gyrB (20\% missing)  \\
\includegraphics[width=0.4\textwidth]{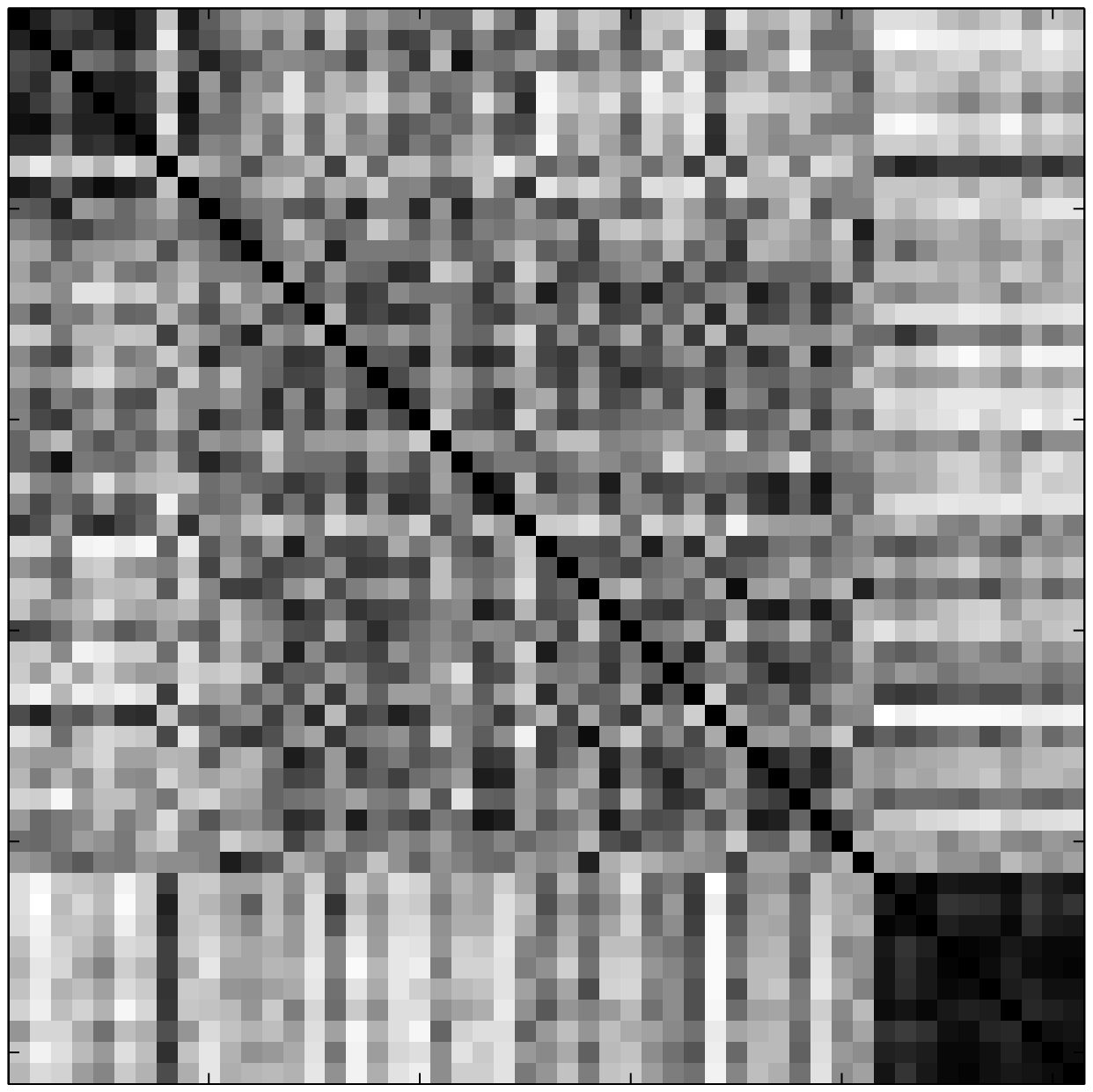} &
\includegraphics[width=0.4\textwidth]{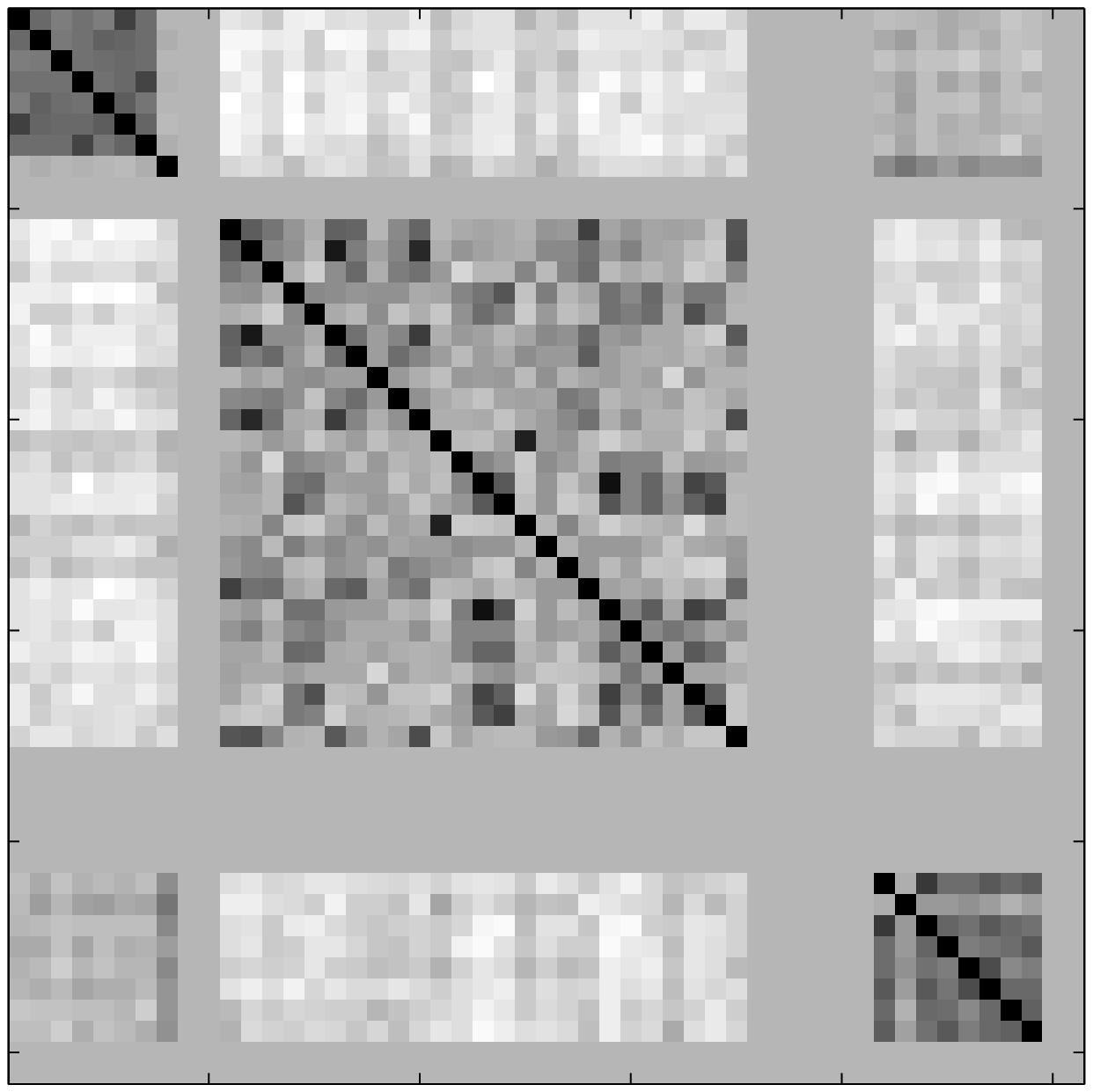} \\
(e) completed matrix & (f) estimated matrix \\
\includegraphics[width=0.4\textwidth]{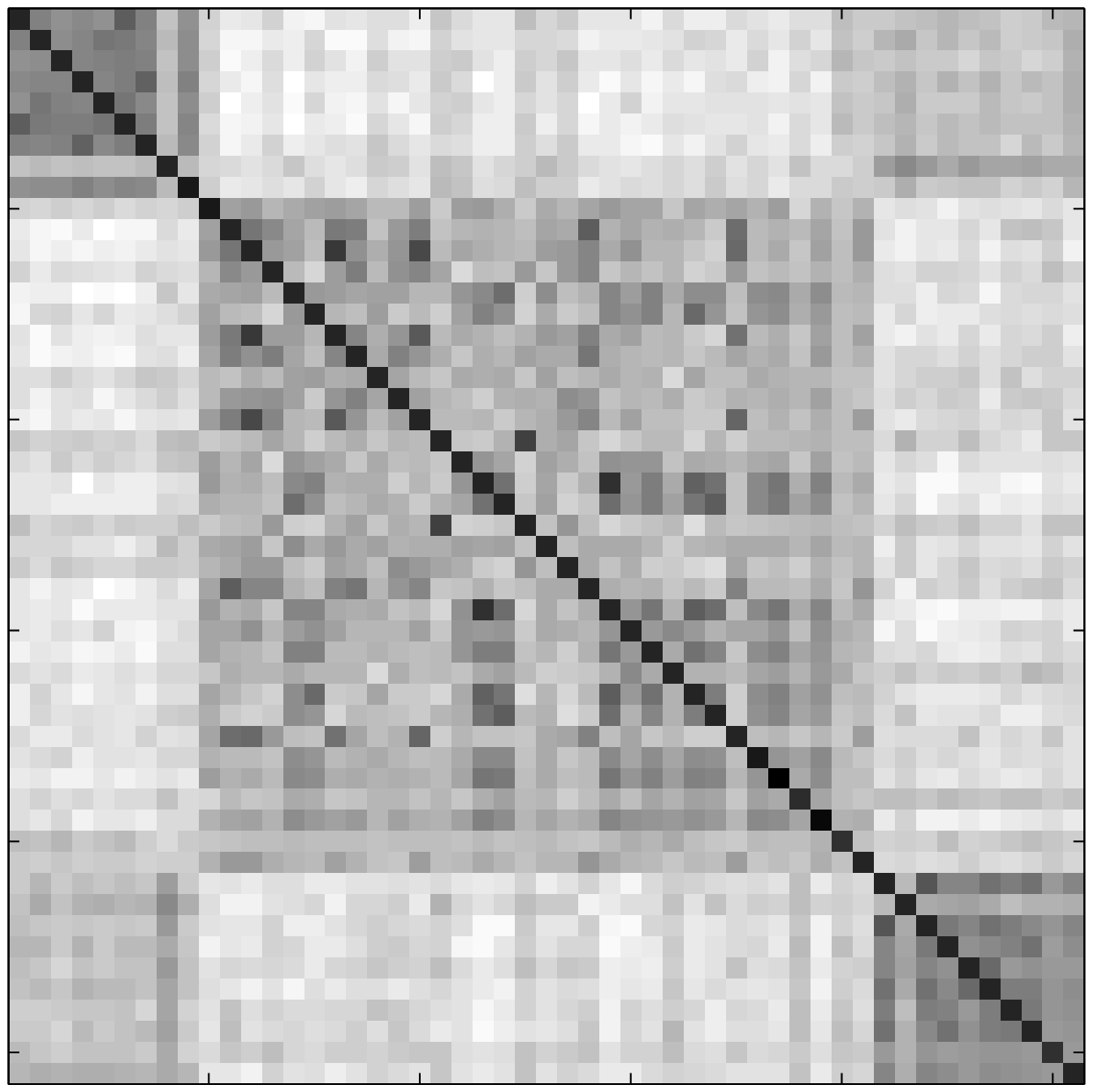} &
\includegraphics[width=0.4\textwidth]{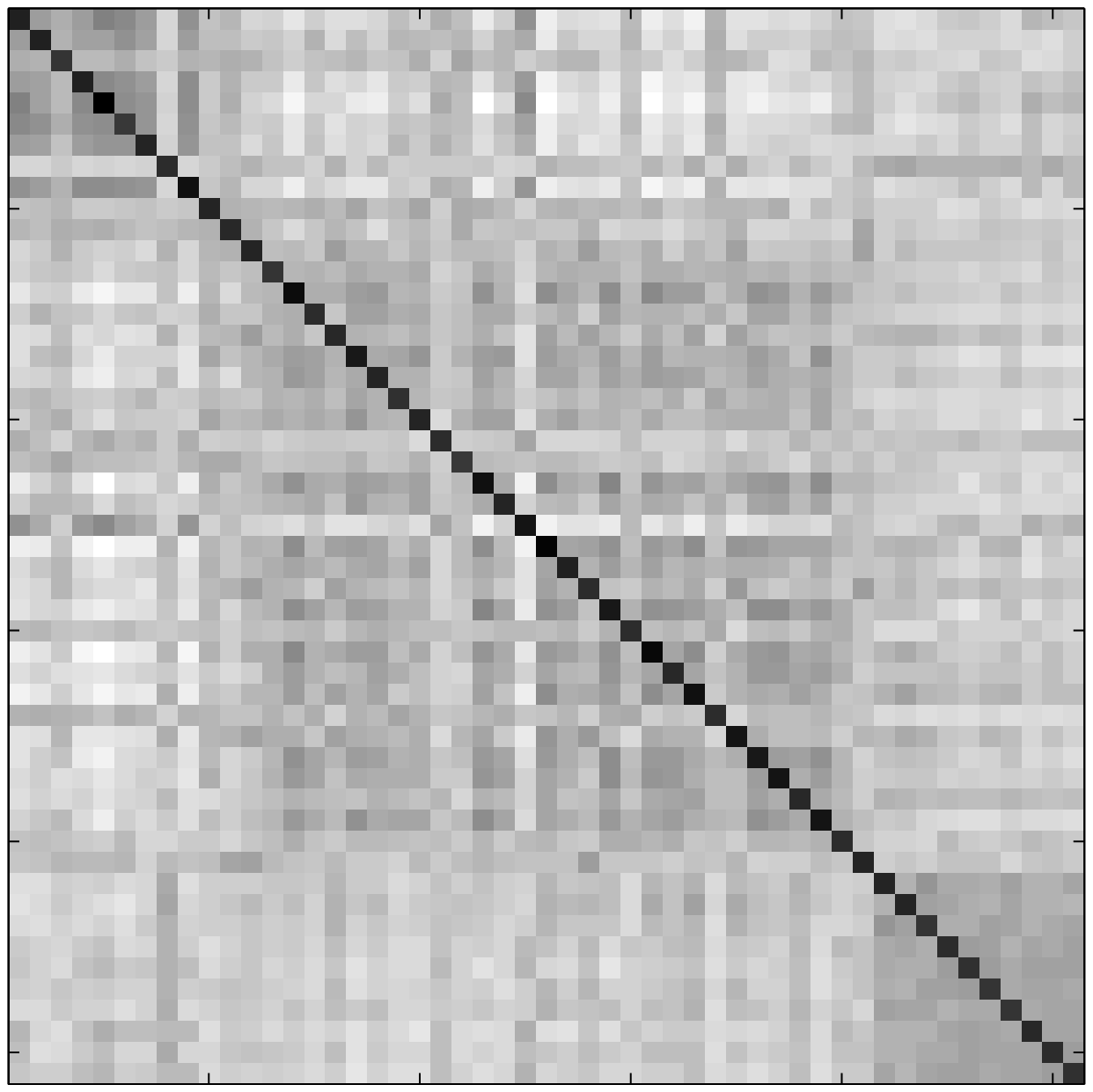} 
\end{tabular}
\end{center}
\caption{An example of kernel matrix completion. See the text for details.}
\label{fig:kermat}
\end{figure}

Our dataset has 52 bacteria of three genera 
({\it Corynebacterium}: 10,
{\it Mycobacterium}: 31,
{\it Rhodococcus}: 11),
each of which has both 16S and gyrB sequences.
For simplicity, let us call these genera as class 1-3, respectively.
For 16S and gyrB, we computed the second order count kernel,
which is the dot product of bimer counts~\citep{TsuKinAsa02}.
Each kernel matrix is normalized such that 
the norm of each sample in the feature space becomes one.
The kernel matrices of gyrB and 16S 
can be seen in Fig.~\ref{fig:kermat} (b) and (c), respectively.
For reference, we show an ideal matrix as Fig.~\ref{fig:kermat}(a), 
which indicates the true classes.
In our senario, for a considerble number of bacteria,
gyrB sequences are not available as in Fig.~\ref{fig:kermat}(d). 
We will complete the missing entries 
by the {\it em} algorithm with the spectral variants of 16S matrix.
When the {\it em} algorithm converges, 
we end up with two matrices: 
the {\it completed matrix} on data manifold ${\cal D}$ 
(Fig.~\ref{fig:kermat}(e))
and the {\it estimated matrix} on model manifold ${\cal M}$
 (Fig.~\ref{fig:kermat}(f)).
These two matrices are in general not the same, 
because the two manifolds may not have intersection.

In order to evaluate the quality of completed and estimated matrices,
K-means clustering is performed in the feature space of each kernel.
In evaluating the partition, we use the Adjusted Rand Index 
(ARI)~\citep{HubAra85,YeuRuz01}. Let $U_1,\ldots,U_c$ be the obtained 
clusters and $T_1,\ldots,T_s$ be the ground truth clusters.
Let $n_{ij}$ be the number of samples which belongs to both $U_i$ and $T_j$.
Also let $n_{i.}$ and $n_{.j}$ be the number of samples in $U_i$ and $T_j$,
respectively.
ARI is defined as
\[
\frac{\sum_{i,j} \bp n_{ij} \\ 2 \ep - \left[ \sum_i \bp n_{i.} \\ 2 \ep
\sum_j \bp n_{.j} 2 \ep \right] / \bp n \\ 2 \ep}
{\frac{1}{2}\left[ \sum_i \bp n_{i.} \\ 2 \ep + \sum_j \bp n_{.j} \\ 2 \ep
\right] - \left[ \sum_i \bp n_{i.} \\ 2 \ep \sum_j \bp n_{.j} \\ 2 \ep \right]
/ \bp n \\ 2 \ep}.
\]
The attractive point of ARI is that it can measure the difference of 
two partitions even when the number of clusters is different.
When the two partitions are exactly the same, ARI is 1, and 
the expected value of ARI over random partitions is 0 
(see \citet{HubAra85} for details).

\begin{figure}
\begin{center}
\includegraphics[width=0.6\textwidth]{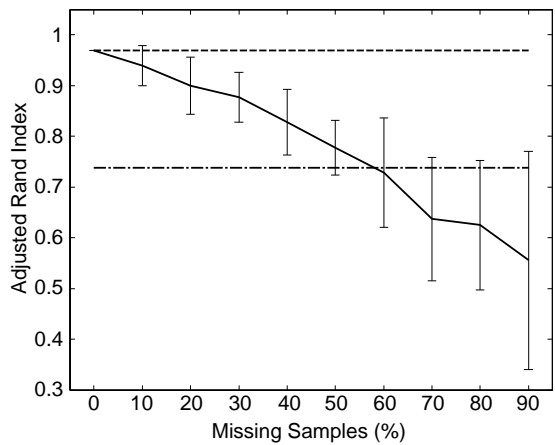}
\end{center}
\caption{Clustering performance of the completed matrix. 
The solid curve shows the averaged ARI of the completed matrix, 
and the error bar describes the standard deviation.
The upper and lower flat lines show the ARIs of the complete gyrB and 16S kernel matrices, respectively.}
\label{fig:ari_completed}
\end{figure}

\begin{figure}
\begin{center}
\includegraphics[width=0.6\textwidth]{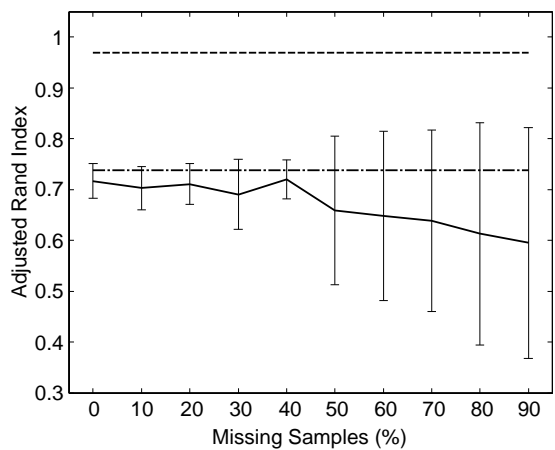}
\end{center}
\caption{Clustering performance of the estimated matrix.
The solid curve shows the averaged ARI of the estimated matrix, 
and the error bar describes the standard deviation.}
\label{fig:ari_estimated}
\end{figure}

The clustering experiment is performed by randomly 
removing samples from gyrB data. 
The ratio of missing samples is changed from 0\% to 90\%.
The ARIs of completed and estimated matrices 
averaged over 20 trials are shown in Fig.~\ref{fig:ari_completed} 
and \ref{fig:ari_estimated}, respectively.
Comparing the two matrices, 
the estimated matrix performed significantly worse than the complete matrix.
It is because the completed matrix maintains existing entries unchanged,
and so the class information in gyrB matrix is well preserved.
We especially focus on the comparison between the completed matrix 
and 16S matrix, 
because there is no point in performing the {\it em} algorithm
when 16S matrix works better than the completed matrix.
According to the plot, the ARI of completed matrix 
was larger than 16S matrix up to 50\% missing ratio.
It implies that the matrix completion is meaningful 
even in quite hard situations ---
50\% sample loss implies 75\% loss in entries.
This result encourages us (and hopefully readers) to apply 
the {\it em} algorithm to other data such as gene networks~\citep{VerKan02}.

\section{Possible Extension} \label{sec:ext}
As we related the {\it em} algorithm to maximum likelihood inference 
in Sec.~\ref{sec:EM}, 
it is straightforward to generalize it to the maximum {\it a posteriori}
(MAP) inference or more generally the Bayes inference~\citep{Cri94}.
For example, we are going to modify the {\it em} algorithm 
to obtain the MAP estimate.
The MAP estimation amounts to minimizing the KL divergence
penalized by a prior,
\[
  KL(D,M) - \log \pi(M), \qquad D\in {\cal D}, \quad M \in {\cal M},
\]
where $\pi(M)$ is a prior distribution for $M$.
Since the additional term $-\log\pi(M)$ depends only on the model $M$,
only the $m$-step is changed so as to minimize the above objective
function with respect to $M$.

Let us give a simple example of MAP estimation in
the spectral variants case.
In Bayesian inference, it is common to take a {\it conjugate prior}, 
so that the posterior distribution remains as a member of
the exponential family.
Since the model parameter $\b$ is related to a covariance matrix,
we choose the Gamma distribution, which works as a conjugate prior 
for the variance of Gaussian distribution~\citep{Cri94}.
The prior distribution is defined independently for each $b_j$ as
\[
 \pi(b_j;\,\nu,\alpha)=\frac{1}{\Gamma(\nu)\alpha^\nu}
 \exp\left\{-\frac{b_j}{\alpha}+(\nu-1)\log b_j\right\},
\]
where $\nu$ and $\alpha$ denote hyperparameters, by which
the mean and the variance are specified by
$E(b_j)=\alpha\nu$ and $V(b_j)=\alpha^2\nu$.
The $m$ step for MAP estimation is to minimize
\[
L_m^{\rm MAP} = L_m - \sum_{j=1}^\ell \log\pi(b_j;\,\nu,\alpha),
\]
which leads to the equation
\[
 \tr(M_i (\sum_{j=1}^c b_j M_j)^{-1}) +\frac{\nu-1}{b_i}
 = \tr(M_i D) + \frac{1}{\alpha},
 \qquad i = 1,\ldots,\ell.
\]
In the spectral variants case, the left hand side is reduced to
$\nu/b_i$, thus we obtain the MAP solution in a closed form as
\[
 b_i = \frac{\nu}{\tr(M_iD)+ 1/\alpha},
 \qquad i = 1,\ldots,\ell.
\]

\section{Conclusion} \label{sec:con}
In this paper, we introduced the information geometry 
in the space of kernel matrices, and applied the {\it em} algorithm
in matrix approximation.
The main difference to other Euclidean methods is 
that we use the KL divergence.
In general, we cannot determine which distance is better, 
because it is highly data dependent.
However our method has a great utility, 
because it can be implemented only with algebraic computation
and we do not need any specialized optimizer such as semidefinite programmming 
unlike \citep{Gra02,CriShaKanEli02,LanCriBarElGJor02}.

One of our contribution is that we related matrix approximation
to statistical inference in Sec.~\ref{sec:EM}.
Thus, in future works, it would be interesting to involve 
advanced methods in statistical inference, 
such as generalized EM~\citep{DemLaiRub77} and variational Bayes~\citep{Att99}.
Also we are looking forward to apply our method to diverse kinds 
of real data which are not limited to bioinformatics.

\appendix
\section{Analytical Solvability of the $m$-step}

In this appendix, we discuss the solvability of the $m$-step.
The left hand side of (\ref{eq:mstep}) is the
$m$-coordinate of the submanifold $\cal M$, while $b_j$ denote
the $e$-coordinate of $\cal M$. The $e$-coordinate and $m$-coordinate
are connected by the Legendre transform~\citep{AmaNag01}.
In the mother manifold ${\cal P}$, the Legendre transform is easily
obtained as the inverse of the matrix. In the submanifold
$\cal M$ of $\cal P$, however, it is difficult to obtain the Legendre
transform in general. The difficulty is caused by the difference of 
geodesics defined in $\cal M$ and $\cal P$.
When the geodesic defined by a coordinate system of a
submanifold ${\cal S} \subseteq {\cal P}$ 
coincides the geodesic defined by the corresponding global 
coordinate system of ${\cal P}$,
the submanifold is called {\it autoparallel}.
In our case, $\cal M$ is autoparallel for the $e$-coordinate,
but it is not always autoparallel for the $m$-coordinate.
When the submanifold is autoparallel for the both coordinate systems,
the submanifold is called doubly autoparallel.

%
%
Let us consider when a submanifold becomes doubly autoparallel.
To begin with, let us define the product $*$ between two $d \times d$ 
symmetric matrices $X, Y\in {\it Sym}(d)$,
\begin{equation} \label{eq:prod}
 X * Y = \frac{1}{2}(X Y + Y X).
\end{equation}
The algebra equipped with the usual matrix sum and the product (\ref{eq:prod})
is called the Jordan algebra of the vector space of {\it Sym}$(d)$.
The following theorem provides the necessary and sufficient condition
for doubly autoparallel submanifold.
\begin{theorem}[\citet{Oha98}, Theorem 4.6]
 Assume the identity matrix $I$ is an element of the submanifold $\cal M$,
 Then $\cal M$ is doubly autoparallel if and only if the tangent space
 of $\cal M$ is a Jordan subalgebra of {\it Sym}$(d)$.
\end{theorem}
When a submanifold ${\cal M} \subseteq {\cal P}$ is determined as 
(\ref{eq:generalM}),
${\cal M}$ is doubly autoparallel if the following holds for all $i, j$:
 \[
  N_i * N_j \in \mathrm{span}(\{N_1,\ldots,N_c\}).
 \]
\citet{Oha98} has shown that, if and only if ${\cal M}$ is doubly autoparallel,
the $m$-projection can be solved analytically, 
that is, the optimal solution is obtained by one Newton step.
For example, in the spectral variants case, $N_i = \v_i \v_i^\top$ and
\[
N_i * N_j = 0 \in \mathrm{span}(\{N_1,\ldots,N_c\}).
\]
Thus the $m$-projection is obtained analytically in this case.

\section*{Acknowledgement}
The authors gratefully acknowledge that 
the bacterial {\it gyrB} amino acid sequences are
offered by courtesy of Identification and Classification of Bacteria (ICB)
database team of Marine Biotechnology Institute, Kamaishi, Japan.
The authors would like to thank T.~Kin, Y.~Nishimori, T.~Tsuchiya and J.-P.~Vert for fruitful discussions.


\end{document}